\documentclass[runningheads]{llncs}
\usepackage[T1]{fontenc}
\usepackage{graphicx}
\usepackage{amsfonts}
\usepackage{amssymb}
\usepackage{marvosym}
\usepackage{booktabs}
\usepackage{array}
\usepackage{multirow}
\usepackage{subcaption}
\usepackage{amsmath}
\usepackage{float}
\usepackage[hidelinks]{hyperref}
\usepackage{orcidlink}
\begin{document}
\title{Recovering Diagnostic Value: Super-Resolution–Aided Echocardiographic Classification in Resource-Constrained Imaging}
\titlerunning{Super-Resolution for Resource-Constrained Echocardiographic Classification}
%

\author{Krishan Agyakari Raja Babu\inst{1}\textsuperscript{(\Letter)}\orcidlink{0000-0001-9362-2420} 
\and
Om Prabhu\inst{2}\orcidlink{0009-0000-9568-3067} 
\and
Annu\inst{3}\orcidlink{0000-0002-2212-6806} 
\and
Mohanasankar Sivaprakasam\inst{1}\orcidlink{0000-0002-6714-9147} 
}

\institute{Indian Institute of Technology Madras, Chennai, India \\ \email{ 
rajababu@alumni.iitm.ac.in} \and All India Institute of Medical Sciences, New Delhi, India \and Indian Institute of Technology Hyderabad, Sangareddy, India
}

\authorrunning{K.A. Raja Babu et al.}
%
\maketitle              
\begin{abstract}
Automated cardiac interpretation in resource-constrained settings (RCS) is often hindered by poor-quality echocardiographic imaging, limiting the effectiveness of downstream diagnostic models. While super-resolution (SR) techniques have shown promise in enhancing magnetic resonance imaging (MRI) and computed tomography (CT) scans, their application to echocardiography—a widely accessible but noise-prone modality—remains underexplored. In this work, we investigate the potential of deep learning–based SR to improve classification accuracy on low-quality 2D echocardiograms. Using the publicly available CAMUS dataset, we stratify samples by image quality and evaluate two clinically relevant tasks of varying complexity: a relatively simple Two-Chamber vs. Four-Chamber (2CH vs. 4CH) view classification and a more complex End-Diastole vs. End-Systole (ED vs. ES) phase classification. We apply two widely used SR models—Super-Resolution Generative Adversarial Network (SRGAN) and Super-Resolution Residual Network (SRResNet), to enhance poor-quality images and observe significant gains in performance metric—particularly with SRResNet, which also offers computational efficiency. Our findings demonstrate that SR can effectively recover diagnostic value in degraded echo scans, making it a viable tool for AI-assisted care in RCS, achieving more with less.


\end{abstract}
%
%
\section{Introduction}
\label{sec:intro}
Echocardiography is one of the most widely used cardiac imaging modalities, valued for its real-time capability, portability, and affordability. These attributes make it particularly critical in resource-constrained settings (RCS), including rural clinics and low- and middle-income countries (LMICs) \cite{FRIJA2021101034,Hamza2024AIEcho,Marangou2019Echocardiography}. However, the diagnostic utility of echocardiography is often undermined by poor image quality. Studies have reported that echocardiographic scans performed with handheld or low-end devices in LMICs are frequently suboptimal for clinical interpretation \cite{Becker2016Portable,ChamsiPasha2017Handheld}. This is primarily due to factors such as limited imaging hardware, variability in operator expertise, and difficult acquisition conditions (e.g., in emergency or bedside scenarios) \cite{Michelis2019,Vedanthan2014Bioimaging,Wierda2023HandheldEcho}.

Poor-quality echo scans not only hinder human interpretation but also significantly degrade the performance of automated tools for view classification, chamber quantification, and disease prediction\textemdash technologies that are increasingly being deployed to address the shortage of trained specialists in such regions \cite{Akkus2021AI_Echo,Hirata2025AI_Echo,Xochicale2023MachineLearningCaseStudy}. Despite the growing reliance on AI-assisted diagnostic pipelines, relatively few studies address the pre-processing bottleneck of enhancing low-fidelity echocardiographic inputs to improve downstream performance \cite{Ashrafian2024VisionLanguageSyntheticDataEnhances,Fernandez2024DiffusionEchocardiography}.

Super-resolution (SR) has emerged as a promising solution for enhancing the quality of medical images, particularly in scenarios where high-resolution acquisition is limited by hardware constraints. While SR techniques have achieved notable success in high-contrast modalities such as magnetic resonance imaging (MRI) and computed tomography (CT) \cite{Isaac2015SuperResolution,Shin2024SuperResolution}, their application to echocardiography—arguably one of the most noise-prone and variable modalities—remains limited. The combination of speckle noise, inconsistent probe positioning, and patient-dependent acoustic windows makes SR in echocardiography a uniquely challenging problem \cite{AbdelNasser2017Enhancement,Gifani2016TSR}. Moreover, most downstream AI models assume access to high-quality input images, overlooking a critical bottleneck in RCS, where degraded image quality is often the norm rather than the exception \cite{Hamza2024AIEcho,Xochicale2023MachineLearningCaseStudy}.

\vspace{1em}
\noindent\textbf{Our contributions are summarized as follows:}
\begin{itemize}
\item We investigate the underexplored application of super-resolution in 2D echocardiography and position it as a lightweight, model-agnostic pre-processing step for enhancing AI-based clinical interpretation in RCS.
\item We demonstrate that super-resolution significantly improves classification performance on degraded scans, with SRResNet offering a favorable trade-off between diagnostic accuracy and computational efficiency.
\end{itemize}
\section{Related Work}
\label{sec:related_work}
SR techniques in echocardiography have evolved from traditional signal processing approaches to recent deep learning-based models. Early efforts predominantly explored temporal super-resolution, aiming to improve the frame rate of echocardiographic videos. Gifani et al. \cite{Gifani2016TSR} introduced a sparse representation-based method that reconstructed intermediate frames using learned dictionaries. Similarly, Afrakhteh et al. \cite{Afrakhteh2022TemporalSR} proposed a high-precision interpolation technique leveraging non-polynomial functions to enhance temporal continuity in ultrasound sequences. While effective in improving temporal resolution, these methods offered limited enhancement in spatial fidelity—an equally critical factor for diagnostic interpretation.

The development of deep learning methods has opened new avenues for spatial SR in ultrasound imaging. Abdel-Nasser and Omer \cite{AbdelNasser2017Enhancement} applied a CNN-based architecture for general ultrasound image enhancement, reporting improvements in structural details and contrast. Cammarasana et al. \cite{Cammarasana2023SuperResolution} proposed a patch-based SR method for 2D ultrasound images and videos, demonstrating gains in spatial resolution across generic anatomical scenes. Similarly, Li et al. \cite{9630440} integrated speckle reduction with deep learning-based SR to enhance segmentation performance in ultrasonic echo images. Unlike prior efforts that primarily focus on perceptual or technical image quality, our study assesses how SR translates into clinically meaningful improvements—especially under RCS.

\section{Methodology}
\label{sec:method}
\begin{figure}[ht]
    \centering
    \resizebox{0.65\linewidth}{!}{
        \includegraphics{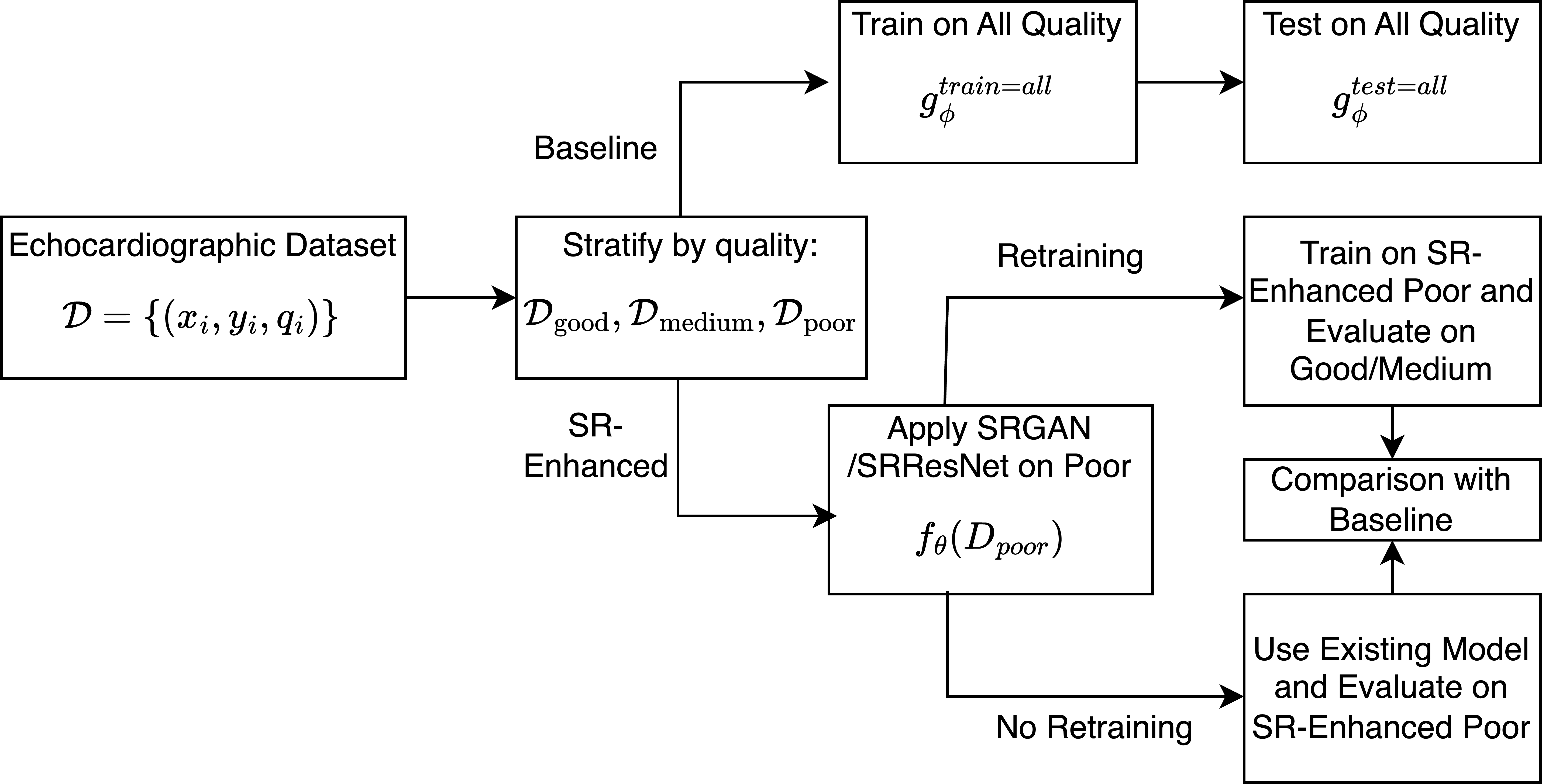}
    }
    \caption{Proposed workflow for super-resolution–aided echo-classification.}
    \label{fig:sr_pipeline}
\end{figure}

We adopt a two-stage pipeline, as illustrated in Fig.~\ref{fig:sr_pipeline}, to assess the diagnostic utility of SR in enhancing low-quality echocardiographic images. Let $\mathcal{D} = \{(x_i, y_i, q_i)\}_{i=1}^N$ denote the dataset, where $x_i \in \mathbb{R}^{H \times W}$ is a 2D echocardiographic frame, $y_i \in \mathcal{Y}$ is the corresponding diagnostic label (e.g., view type or cardiac phase), and $q_i \in \{\text{good}, \text{medium}, \text{poor}\}$ is the image quality metadata provided by clinical experts. Rather than using pixel resolution as a proxy for image quality—which is often misleading—we utilize this clinically validated $q_i$ to stratify the dataset into three disjoint subsets: $\mathcal{D}_{\text{good}}$, $\mathcal{D}_{\text{medium}}$, and $\mathcal{D}_{\text{poor}}$.

We treat $\mathcal{D}_{\text{poor}}$ as a representative of images acquired in RCS. To assess diagnostic performance across varying image quality, we define a classifier $g_{\phi}: \mathbb{R}^{H \times W} \rightarrow \mathcal{Y}$ trained independently on each subset and tested across all quality levels. We consider two clinically relevant classification tasks: (i) a simple view classification of two-chamber vs. four-chamber (2CH vs. 4CH) and (ii) a complex cardiac phase classification of End-Diastole vs. End-Systole (ED vs. ES). The goal is to observe how image quality affects $g_\phi$'s ability to extract diagnostic features without spurious biases.

For enhancement, we introduce a super-resolution module $f_{\theta}: \mathbb{R}^{H \times W} \rightarrow \mathbb{R}^{rH \times rW}$, where $r$ is the upsampling factor. We investigate two SR architectures: SRGAN and SRResNet, pretrained on natural images and fine-tuned on echocardiographic images. The SR-enhanced poor quality subset is then denoted as $\mathcal{D}_{\text{poor}}^{\text{SR}} = \{(f_\theta(x_i), y_i) \mid (x_i, y_i) \in \mathcal{D}_{\text{poor}}\}$. We re-evaluate $g_\phi$ using $\mathcal{D}_{\text{poor}}^{\text{SR}}$ to quantify the gains in classification metric post enhancement. This approach allows us to systematically evaluate the extent to which super-resolution can recover diagnostic information from clinically degraded echocardiograms and improve the utility of downstream AI models in RCS.

\section{Experiments}
\label{sec:expt}
\subsection{Dataset and Quality Stratification}

We conduct our experiments on the publicly available "Cardiac Acquisitions for Multi-structure Ultrasound Segmentation" (CAMUS) dataset \cite{camus}, which consists of 2D echocardiographic sequences from 500 patients. For each patient, ED and ES frames are annotated in both 2CH and 4CH views, yielding a total of 2,000 labeled frames. Importantly, the dataset includes expert-provided image quality annotations—categorized as good, medium, or poor—which we leverage for clinically grounded stratification. Unlike resolution-based proxies, this metadata reflects real-world diagnostic usability, making it more relevant for evaluating model performance in practical settings.

Table~\ref{tab:image_quality_distribution} summarizes the distribution of echo images across views, cardiac phases and image quality in the dataset. In the 2CH view, 43.4\% of frames are rated as good, 42.8\% as medium, and 13.8\% as poor. Similarly, in the 4CH view, 57.6\% of frames are rated as good, 33.0\% as medium, and 9.4\% as poor. While most frames fall into the good or medium categories, a clinically meaningful subset—232 poor-quality frames—reflects imaging conditions typical of RCS, and thus serves as our primary focus for enhancement and evaluation.

\begin{figure}[H]
\centering
\setlength{\tabcolsep}{3pt}
\renewcommand{\arraystretch}{1.2}
\resizebox{0.93\linewidth}{!}{%
\begin{tabular}{cccccc}
\textbf{2CH-Good-ED} & \textbf{2CH-Medium-ED} & \textbf{2CH-Poor-ED} & \textbf{4CH-Good-ED} & \textbf{4CH-Medium-ED} & \textbf{4CH-Poor-ED} \\
\includegraphics[width=2.5cm,height=2.5cm]{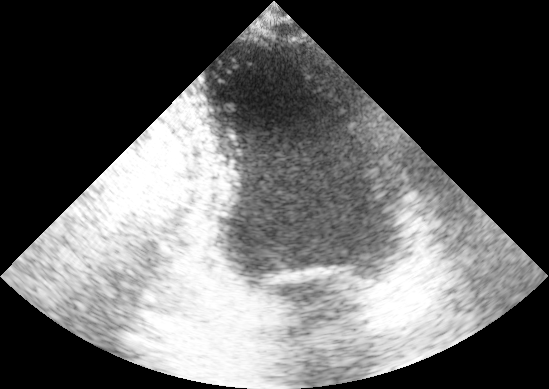} &
\includegraphics[width=2.5cm,height=2.5cm]{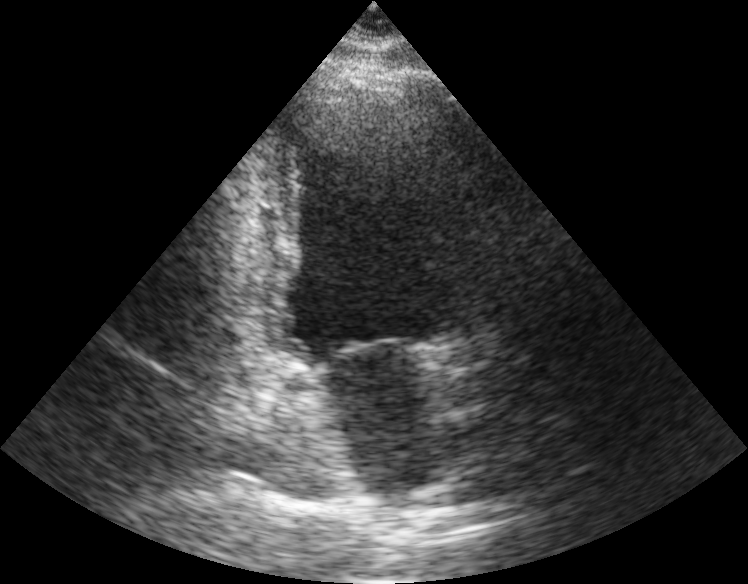} &
\includegraphics[width=2.5cm,height=2.5cm]{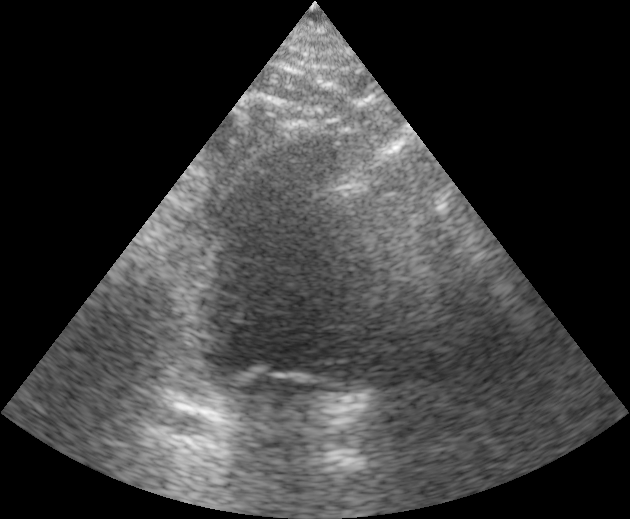} &
\includegraphics[width=2.5cm,height=2.5cm]{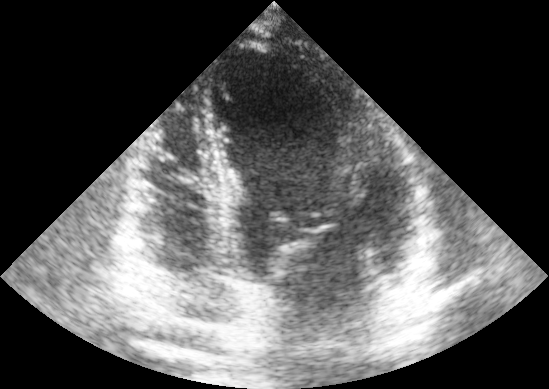} &
\includegraphics[width=2.5cm,height=2.5cm]{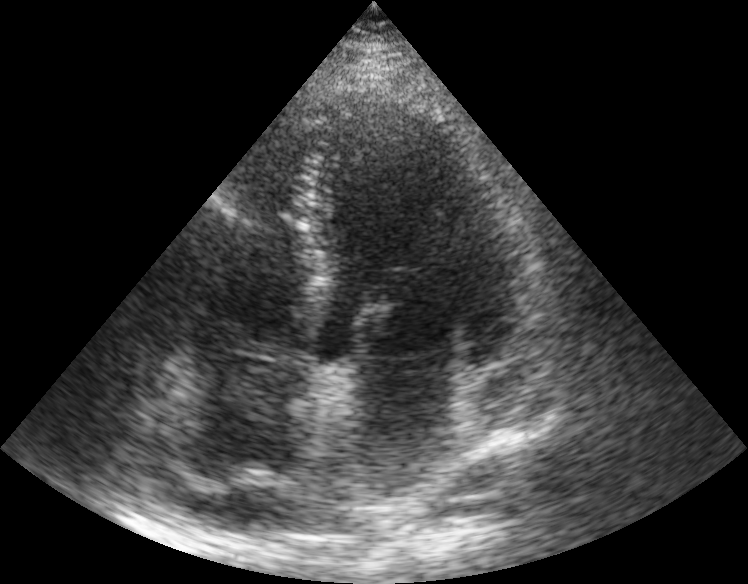} &
\includegraphics[width=2.5cm,height=2.5cm]{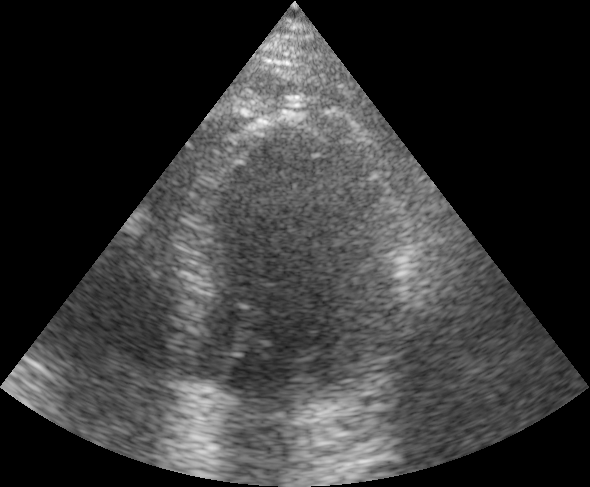} \\
\textbf{2CH-Good-ES} & \textbf{2CH-Medium-ES} & \textbf{2CH-Poor-ES} & \textbf{4CH-Good-ES} & \textbf{4CH-Medium-ES} & \textbf{4CH-Poor-ES} \\
\includegraphics[width=2.5cm,height=2.5cm]{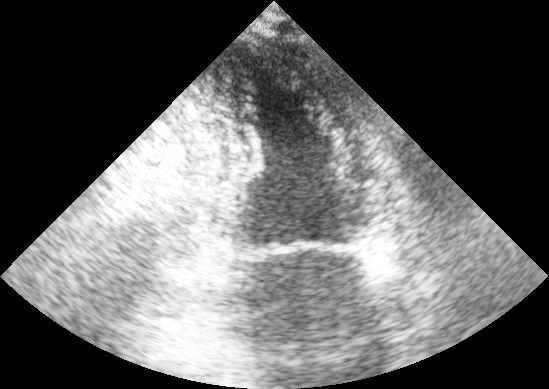} &
\includegraphics[width=2.5cm,height=2.5cm]{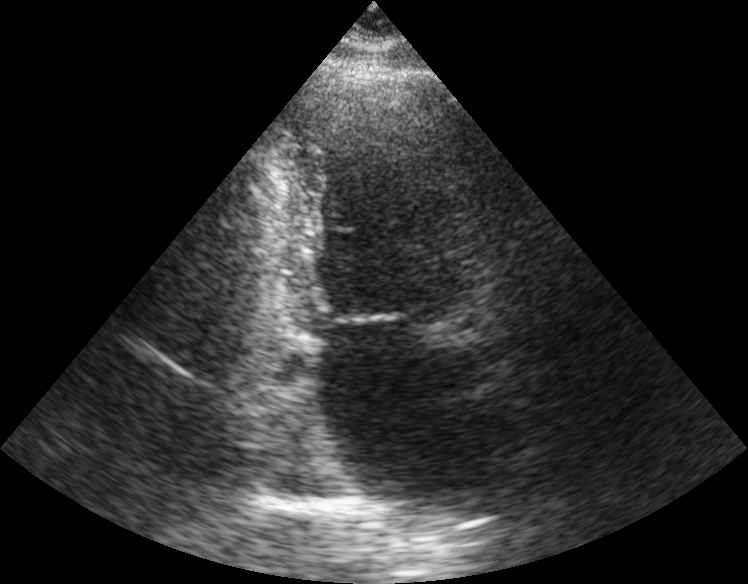} &
\includegraphics[width=2.5cm,height=2.5cm]{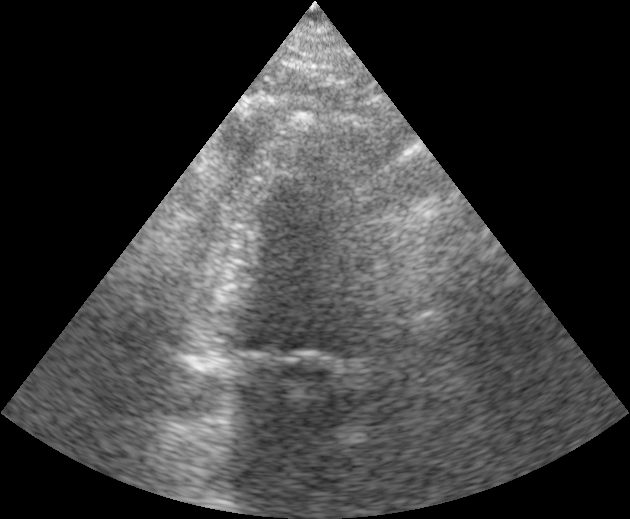} &
\includegraphics[width=2.5cm,height=2.5cm]{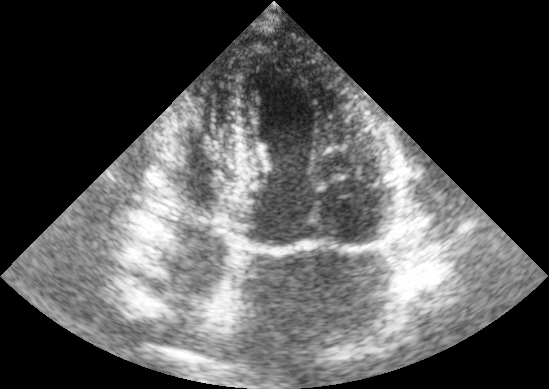} &
\includegraphics[width=2.5cm,height=2.5cm]{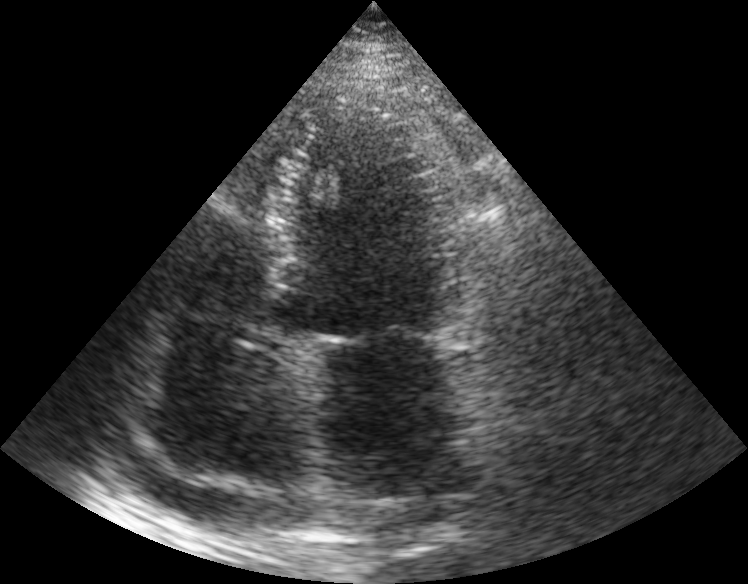} &
\includegraphics[width=2.5cm,height=2.5cm]{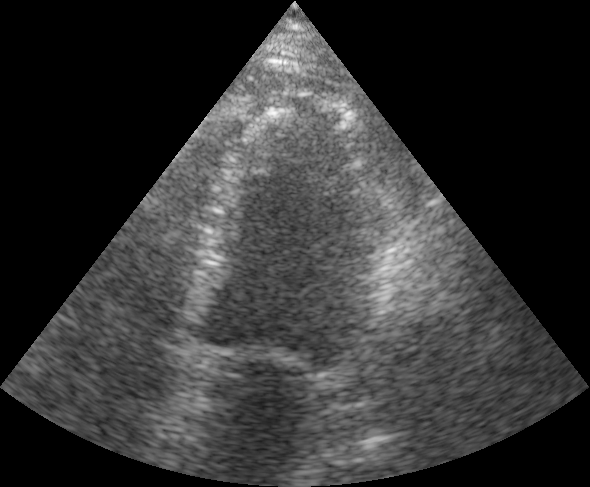} \\
\end{tabular}
}
\caption{Representative echocardiographic frames from the CAMUS dataset across quality levels (Good, Medium, Poor), views (2CH, 4CH), and phases (ED, ES).}
\label{fig:camus_quality_grid}
\end{figure}

Figure~\ref{fig:camus_quality_grid} presents representative echocardiographic frames across the three quality tiers, spanning both views and cardiac phases. As evident from the visual examples,
poor-quality frames exhibit low contrast, speckle noise, and structural ambiguity, posing substantial challenges for both human readers and AI systems. These visual limitations motivate the application of super-resolution as a pre-processing step to enhance diagnostic value under degraded conditions.

\begin{table}[htbp]
\centering
\caption{Image distribution by view, phase, and quality in the CAMUS dataset.}
\label{tab:image_quality_distribution}
 \resizebox{0.5\linewidth}{!}{
\begin{tabular}{@{}llcccc@{}}
\toprule
\textbf{View} & \textbf{Phase} & \textbf{Good} & \textbf{Medium} & \textbf{Poor} & \textbf{Total} \\
\midrule
\multirow{2}{*}{2CH} & ED & 217 & 214 & 69 & 500 \\
                     & ES & 217 & 214 & 69 & 500 \\
\midrule
\multirow{2}{*}{4CH} & ED & 288 & 165 & 47 & 500 \\
                     & ES & 288 & 165 & 47 & 500 \\
\midrule
\textbf{Total} &      & \textbf{1010} & \textbf{758} & \textbf{232} & \textbf{2000} \\
\bottomrule
\end{tabular}
}
\end{table}

\subsection{Classification Tasks and Setup}
To assess the downstream utility of SR, we define two clinically meaningful classification tasks: (i) a simpler view classification task distinguishing between 2CH and 4CH echocardiographic views, and (ii) a more challenging cardiac phase classification task differentiating between ED and ES frames. These tasks serve to evaluate both coarse and fine-grained diagnostic distinctions.

We adopt a ResNet-18 \cite{resnet18} model pretrained on ImageNet as the classification backbone. The model is fine-tuned on echocardiographic images using a batch size of 16 for 10 epochs, with cross-entropy loss as the objective function and the Adam optimizer (learning rate $1 \times 10^{-4}$).

To ensure class balance and control for potential sampling bias, we construct uniformly stratified datasets by taking the number of poor-quality samples as the reference. Specifically, for the 2CH vs 4CH classification task, we use 138 2CH (69 ED + 69 ES) and 94 4CH (47 ED + 47 ES) images. For the ED vs ES classification task, we use 116 ED (69 2CH + 47 4CH) and 116 ES (69 2CH + 47 4CH) images. In both cases, 80\% of the data is used for training and 20\% for testing. This setup allows us to rigorously evaluate the diagnostic discriminative power of images under different quality conditions, and later assess whether super-resolution improves this performance.

\subsection{Super-Resolution Integration}

To train the SR models on domain-specific data, we construct synthetic low-resolution and high-resolution image pairs using the 1,010 good-quality echocardiographic frames from the CAMUS dataset. Each high-quality image is degraded via bicubic downsampling by a factor of 4 to simulate low-resolution inputs, forming paired data for supervised SR training.

We leverage two widely adopted SR architectures—SRResNet and SRGAN \cite{ledig2017photorealisticsingleimagesuperresolution}—both pretrained on the DIV2K dataset \cite{div2k}. SRResNet is a lightweight model trained with pixel-wise mean squared error (MSE) loss, favoring accurate structural reconstruction and fast inference. In contrast, SRGAN employs a more complex generative adversarial framework, combining a perceptual loss with an adversarial loss to generate visually sharper and more realistic outputs. We fine-tune SRResNet for 4,000 epochs and SRGAN for 8,000 epochs on echocardiographic data using a scaling factor of 4.

\begin{figure}[htbp]
    \centering
    \resizebox{0.8\linewidth}{!}{
    \begin{minipage}[b]{0.24\textwidth}
        \centering
        \includegraphics[width=3cm,height=3cm]{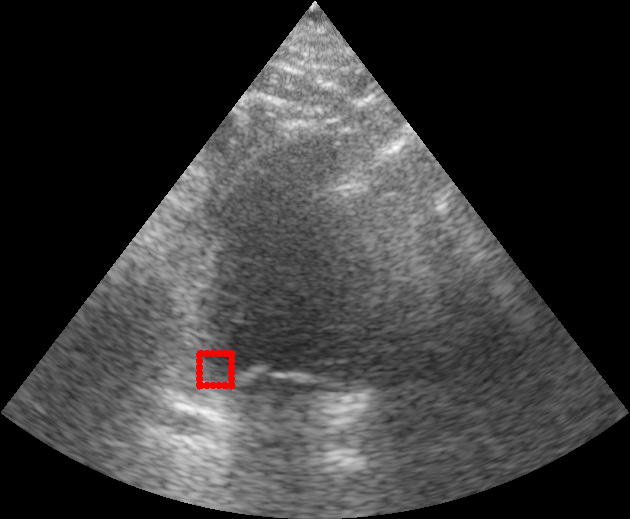}
        \caption*{(a) Reference}
    \end{minipage}
    \hfill
    \begin{minipage}[b]{0.24\textwidth}
        \centering
        \includegraphics[width=3cm,height=3cm]{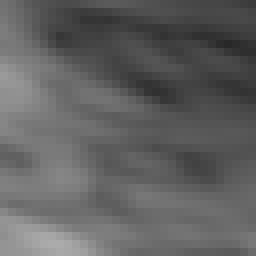}
        \caption*{(b) Cropped Patch}
    \end{minipage}
    \hfill
    \begin{minipage}[b]{0.24\textwidth}
        \centering
        \includegraphics[width=3cm,height=3cm]{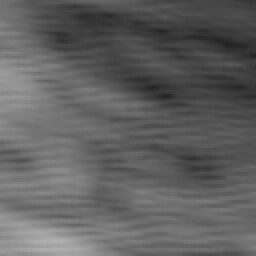}
        \caption*{(c) SRGAN }
    \end{minipage}
    \hfill
    \begin{minipage}[b]{0.24\textwidth}
        \centering
        \includegraphics[width=3cm,height=3cm]{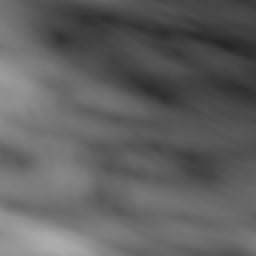}
        \caption*{(d) SRResNet}
    \end{minipage}
    }
    \caption{SR-enhanced outputs for a poor-quality 2CH ED image: (c) SRGAN — 32.70 dB / 0.7164; (d) SRResNet — 38.98 dB / 0.9214 (PSNR / SSIM).}
    \label{fig:magnified_patch_comparison}
\end{figure}

Following training, we apply both models to enhance the 232 poor-quality frames, generating SR-enhanced counterparts. These enhanced images are subsequently used as input for downstream classification tasks to assess whether diagnostic value can be recovered under degraded imaging conditions. A visual comparison of the super-resolved outputs is shown in Figure~\ref{fig:magnified_patch_comparison}, focusing on a representative image patch. Although SRGAN, a perceptual-driven generative model, yields a Peak Signal-to-Noise Ratio (PSNR) of 32.70 dB and a Structural Similarity Index Measure (SSIM) of 0.7164, it fails to recover finer structural details. In contrast, SRResNet achieves notably higher reconstruction fidelity (PSNR = 38.98 dB, SSIM = 0.9214) despite being less computationally demanding—making it a more practical choice for deployment in RCS.

\section{Results and Discussion}
\label{sec:results}
Table~\ref{tab:classification_baseline_table} presents the baseline classification accuracy for both view (2CH vs. 4CH) and phase (ED vs. ES) tasks across all combinations of training and testing image quality, prior to applying SR. Figure \ref{fig:sr_improvement_grid} illustrates the percentage improvement achieved when SR-enhanced poor-quality images are used for training or testing.

\begin{table}[htbp]
\centering
\caption{View and phase classification accuracy(\%) across image quality levels.}
\label{tab:classification_baseline_table}
\setlength{\tabcolsep}{6pt}
\renewcommand{\arraystretch}{1.2}
\resizebox{0.6\linewidth}{!}{%
\begin{tabular}{@{}lccccc@{}}
\toprule
\textbf{Train $\downarrow$ / Test $\rightarrow$} & \multicolumn{1}{c}{\textbf{Quality}} & \textbf{View Accuracy} & \textbf{Phase Accuracy} \\
\midrule
\multirow{3}{*}{\textbf{Good}}   
  & Good   & 100 & 87  \\
  & Medium & 90 & 83  \\
  & Poor   & 82 & 77  \\
\midrule
\multirow{3}{*}{\textbf{Medium}} 
  & Good   & 92 & 85 \\
  & Medium & 100 & 81  \\
  & Poor   & 91 & 81  \\
\midrule
\multirow{3}{*}{\textbf{Poor}}   
  & Good   & 77 & 74  \\
  & Medium &  90& 77 \\
  & Poor   & 100 & 79  \\
\bottomrule
\end{tabular}
}
\end{table}


Based on these results, following observations can be noted:
\begin{enumerate}
    \item \textbf{Diagnostic Collapse: The Cost of Image Degradation Across Tasks and Quality Levels:} As shown in Table~\ref{tab:classification_baseline_table}, in the case of view classification, the model trained on good-quality images performs perfectly on similar data (100\% accuracy), but its accuracy drops sharply—by 10\% on medium and 18\% on poor-quality images. Conversely, the model trained on poor-quality data, while achieving 100\% accuracy on its own domain, experiences a 23\% drop when tested on good images and 10\% on medium. In comparison, phase classification, although it does not reach perfect accuracy in any setting, exhibits lower sensitivity to quality shifts—with an 11.5\% drop from Good→Poor and only 6.3\% from Poor→Good. These findings reveal that image degradation leads to greater diagnostic collapse in the simpler 2CH vs. 4CH view classification—which relies on anatomical structures—compared to the more abstract ED vs. ES phase classification that leverages functional cues. Notably, medium-quality images consistently yield robust performance across test conditions, suggesting they strike an effective balance between noise and structural fidelity—making them an ideal candidate for RCS.

\begin{figure}[htbp]
    \centering

    \begin{minipage}[b]{0.48\textwidth}
        \centering
        \includegraphics[width=4cm,height=3cm]{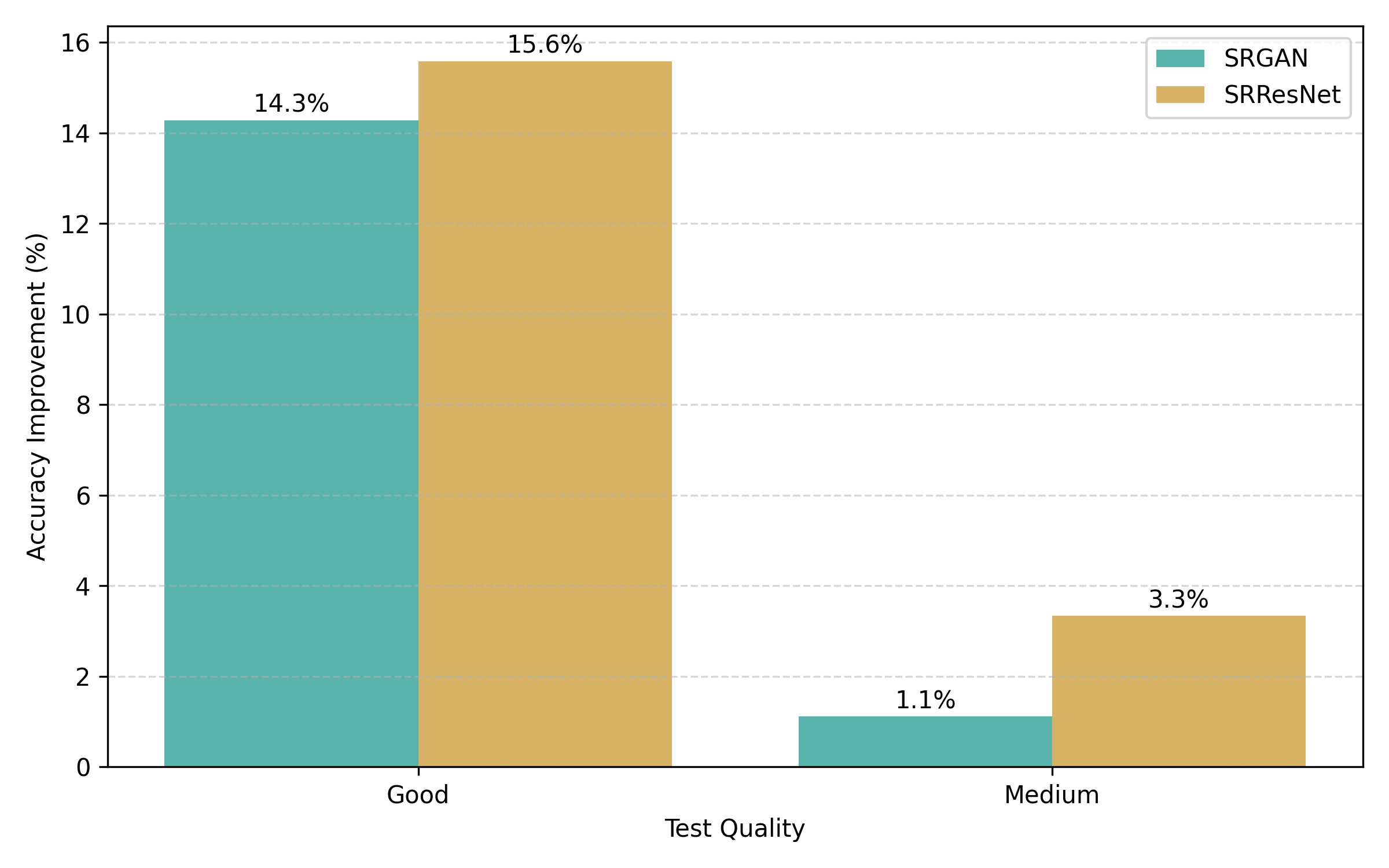}
        \caption*{(a) Train on Poor (2CH vs 4CH)}
    \end{minipage}
    \hfill
    \begin{minipage}[b]{0.48\textwidth}
        \centering
        \includegraphics[width=4cm,height=3cm]{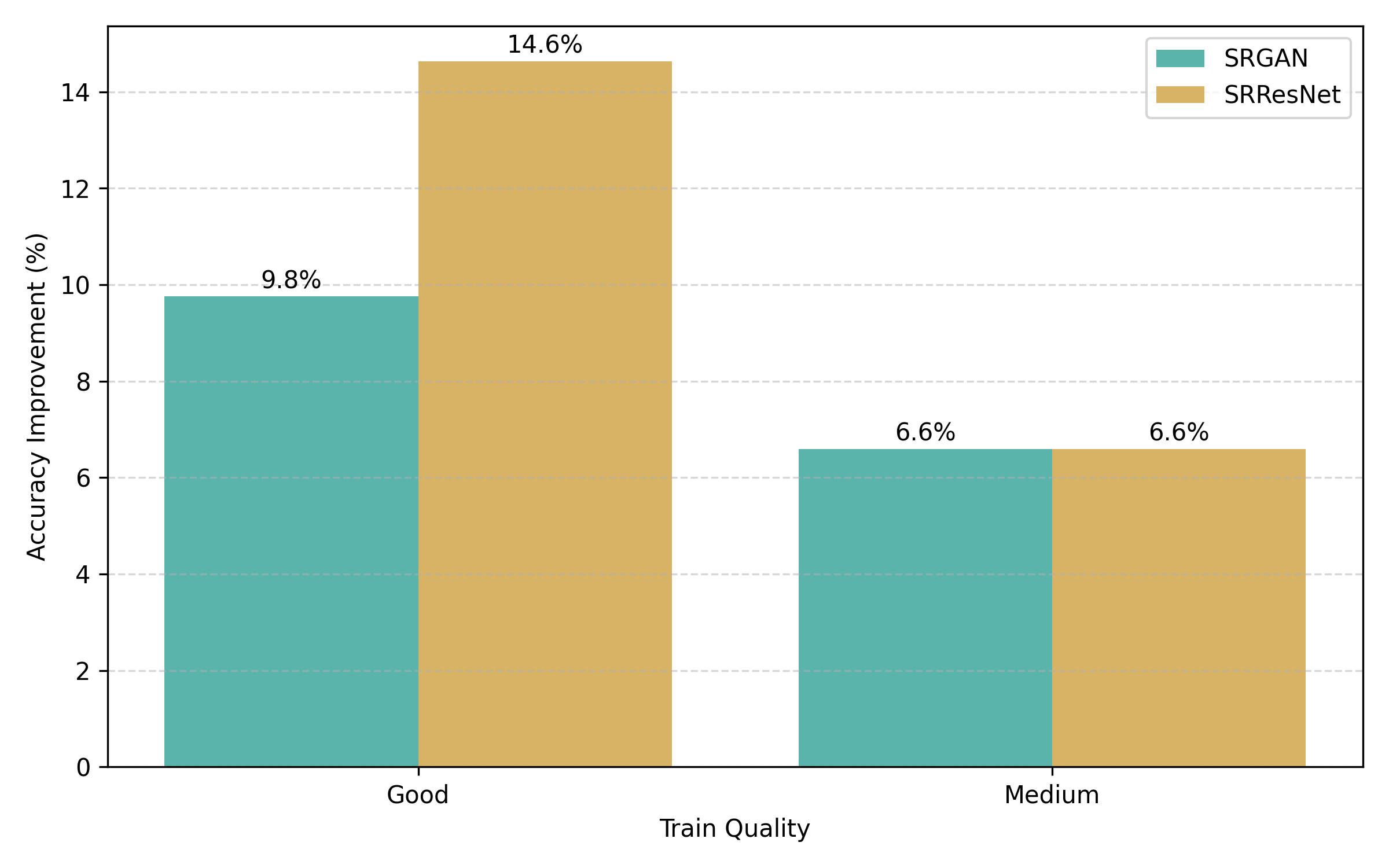}
        \caption*{(b) Test on Poor (2CH vs 4CH)}
    \end{minipage}

    \vspace{1em}

    \begin{minipage}[b]{0.48\textwidth}
        \centering
        \includegraphics[width=4cm,height=3cm]{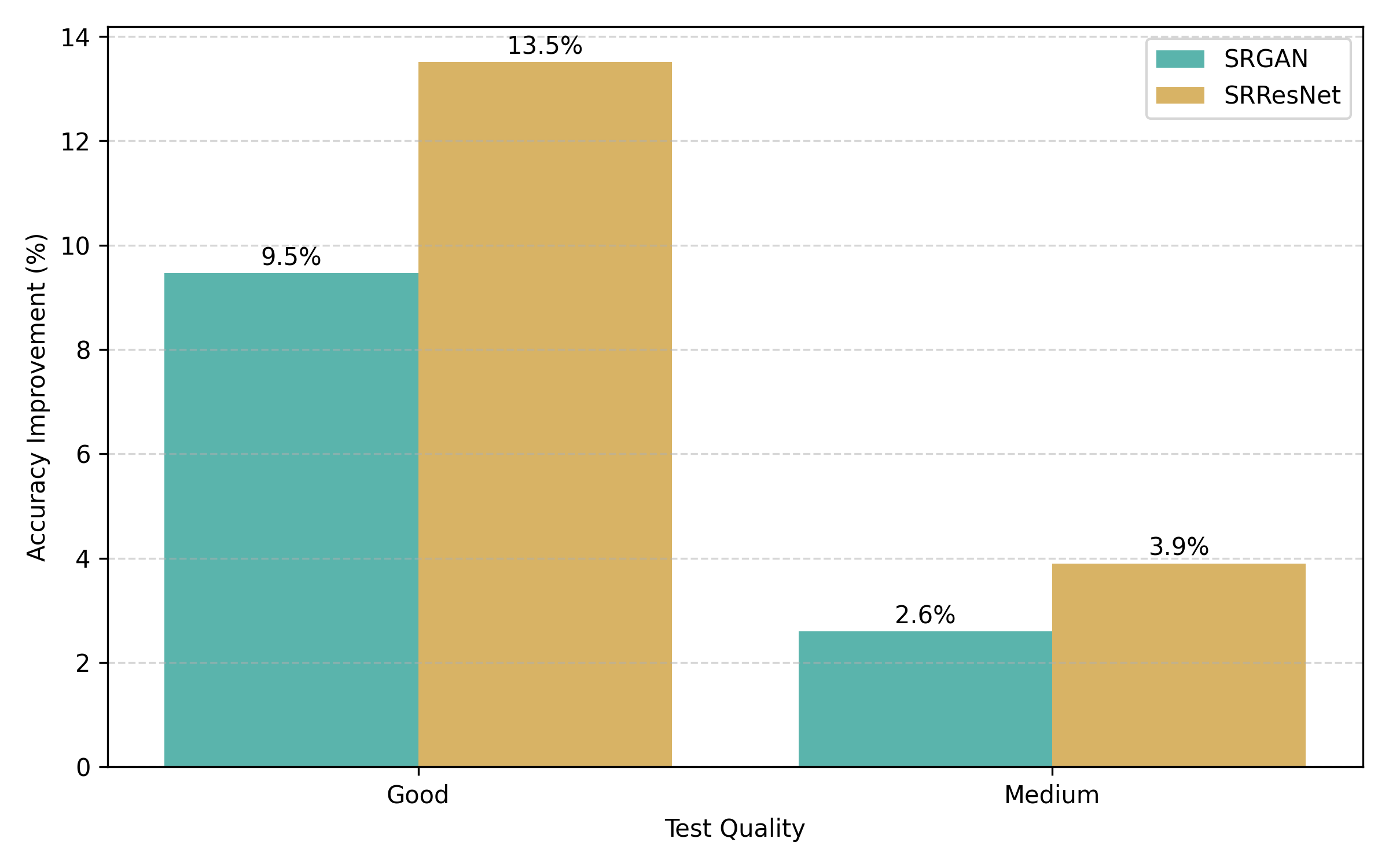}
        \caption*{(c) Train on Poor (ED vs ES)}
    \end{minipage}
    \hfill
    \begin{minipage}[b]{0.48\textwidth}
        \centering
        \includegraphics[width=4cm,height=3cm]{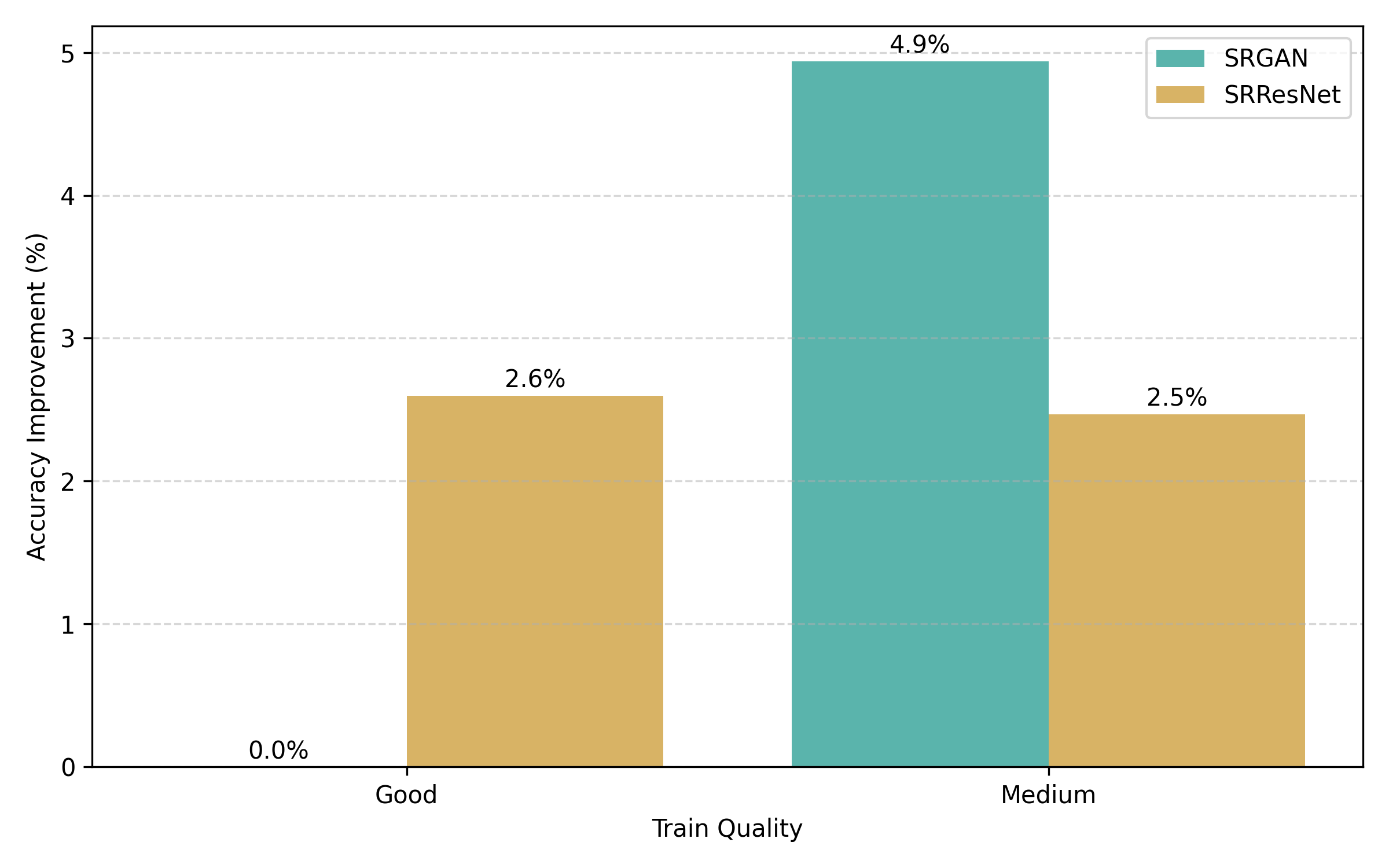}
        \caption*{(d) Test on Poor (ED vs ES)}
    \end{minipage}

    \caption{Percentage improvement in classification performance after integrating super-resolution across two classification tasks (2CH vs 4CH and ED vs ES).}
    \label{fig:sr_improvement_grid}
\end{figure}

    \item \textbf{Recovery Power of Super-Resolution Across Architectures:} As illustrated in Fig.~\ref{fig:sr_improvement_grid}, integrating SR-enhanced poor-quality images into the classification pipeline yields notable gains in both view and phase tasks. As expected, the simpler 2CH vs. 4CH classification benefits more than the functionally complex ED vs. ES task. Interestingly, when model is trained on SR-enhanced poor images and tested on good-quality data, we observe an average accuracy improvement of 14.9\% in the view task and 11.5\% in the phase task—underscoring both the cross-quality generalizability and the restorative potential of SR. Notably, SRResNet—despite being architecturally simpler than the adversarially-trained SRGAN—achieves higher average improvement across all test cases (7.83\% vs. 6.6\%). This suggests that pixel-accurate SR methods like SRResNet may be more suitable for echocardiographic recovery in RCS, where structural fidelity is critical and computational overhead must be minimized.
    
    \item \textbf{Silent Gains with SR-Resolved Evaluation:} Figures~\ref{fig:sr_improvement_grid}(a) and (c) represent scenarios where SR-enhanced poor-quality images are used for re-training, while Figures~\ref{fig:sr_improvement_grid}(b) and (d) show the case where these enhanced images are used solely during evaluation, without modifying the trained models. Remarkably, even without retraining, we observe measurable gains: in the 2CH vs. 4CH task, accuracy improves by an average of 12.2\% when models are trained on good-quality data and by 6.6\% when trained on medium. In the more complex ED vs. ES classification, the improvements are more modest—1.3\% and 3.7\% for models trained on good and medium-quality data, respectively. While these gains may appear subtle, they underscore the practical value of SR as a lightweight, test-time enhancement strategy—particularly beneficial in RCS where retraining may not be feasible.
\end{enumerate}
\section{Conclusion}
\label{sec:conc}
This study highlights the potential of super-resolution (SR) techniques to restore diagnostic utility in degraded echocardiographic images, a frequent limitation in resource-constrained settings. By enhancing poor-quality 2D echo scans with SRGAN and SRResNet, we demonstrate measurable gains in both view and phase classification accuracy—particularly with SRResNet, which consistently outperforms while remaining computationally efficient. Notably, performance improvements are observed even when SR is applied only at inference, underscoring its value as a lightweight preprocessing tool.

Future work should extend this investigation to more complex clinical tasks such as segmentation and disease classification, leveraging larger datasets and more sophisticated deep learning models. Such exploration will be key to validating the broader clinical relevance of SR-enhanced echocardiography and enabling robust AI-assisted diagnostics in low-resource environments.

\subsubsection{\discintname}
The authors have no competing interests to declare that are relevant to the content of this article.
%
%
%
 \bibliographystyle{splncs04}
 \bibliography{mybibliography}
\end{document}